# Covering rough sets based on neighborhoods: An approach without using neighborhoods

Ping Zhu[a,b]

[a]School of Science, Beijing University of Posts and Telecommunications, Beijing 100876, China
[b]State Key Laboratory of Networking and Switching, Beijing University of Posts and Telecommunications, Beijing 100876, China

**Abstract**

Rough set theory, a mathematical tool to deal with inexact or uncertain knowledge in information systems, has originally described the indiscernibility of elements by equivalence relations. Covering rough sets are a natural extension of classical rough sets by relaxing the partitions arising from equivalence relations to coverings. Recently, some topological concepts such as neighborhood have been applied to covering rough sets. In this paper, we further investigate the covering rough sets based on neighborhoods by approximation operations. We show that the upper approximation based on neighborhoods can be defined equivalently without using neighborhoods. To analyze the coverings themselves, we introduce unary and composition operations on coverings. A notion of homomorphism is provided to relate two covering approximation spaces. We also examine the properties of approximations preserved by the operations and homomorphisms, respectively.

## 1. Introduction

Rough set theory, proposed by Pawlak in the early 1980s [24, 25], is a mathematical tool to deal with uncertainty and incomplete information. Since then we have witnessed a systematic, world-wide growth of interest in rough set theory [1, 2, 12, 14, 18, 28, 39, 41, 44, 45, 47, 49, 50, 55, 56, 57, 59] and its applications [4, 11, 17, 26, 27, 30, 48, 58]. Nowadays, it turns out that this approach is of fundamental importance to artificial intelligence and cognitive sciences, especially in the areas of data mining, machine learning, decision analysis, knowledge management, expert systems, and pattern recognition.

Rough set theory bears on the assumption that some elements of a universe may be indiscernible in view of the available information about the elements. Thus, the indiscernibility relation is the starting point of rough set theory. Such a relation was first described by equivalence relation in the way that two elements are related by the relation if and only if they are indiscernible from each other. In this framework, a rough set is a formal approximation of a subset of the universe in terms of a pair of unions of equivalence classes which give the lower and upper approximations of the subset. However, the requirement of equivalence relation as the indiscernibility relation is too restrictive for many applications. In other words, many practical data sets cannot be handled well by classical rough sets. In light of this, equivalence relation has been generalized to characteristic relation [8, 9, 29], similarity relation [34], tolerance relation [5, 6, 23, 32], and even arbitrary binary relation [13, 20, 41, 42, 43, 53] in some extensions of the classical rough sets. Another approach is the relaxation of the partition arising from equivalence relation to a covering. The covering of a universe is used to construct the lower and upper approximations of any subset of the universe [2, 3, 28, 45, 57].

In the literature, several different types of covering-based rough sets have been proposed and investigated; see, for example, [16, 31, 38, 40, 54, 55, 59] and the bibliographies therein. It is well-known that coverings are a fundamental concept in topological spaces and play an important role in the study of topological properties. This motivates the research of covering rough sets from the topology point of view. Some initial attempts have already been made along

*Email address:* pzhubupt@gmail.com (Ping Zhu)

the way. For example, Zhu and Wang examined the topological properties of the lower and upper approximation operations for covering generalized rough sets in [58, 60]. Wu *et al.* combined the notion of topological spaces into rough sets and then discussed the properties of topological rough spaces [37]. In [54], neighborhoods, another elementary concept in topology, have been used to define an upper approximation; some properties of approximation operations for this type of covering rough sets have been explored as well [19, 31, 54, 56].

The purpose of this paper is to investigate further the covering rough sets based on neighborhoods introduced first in [54]. With a little surprise, we find that the same upper approximation as in [54] can be defined without using neighborhoods. (Certainly, it does not mean that the notion of neighborhoods is worthless, as we will see later.) To analyze the coverings for covering-based rough sets, we introduce two operations that allow us to combine, or compose, two or more coverings, as well as several operations on a single covering to modify appropriately the elements of the covering. In order to relate two covering approximation spaces, the notion of homomorphism used extensively in algebra and topology is introduced to covering rough sets. We also examine the properties of the lower and upper approximations preserved by the operations and homomorphisms, respectively.

The remainder of this paper is structured as follows. In Section 2, we briefly review some basics of covering rough sets based on neighborhoods and provide a new equivalent characterization of the upper approximation without using neighborhoods. Based upon the equivalent characterization, the relationships between the lower and upper approximations are discussed in this section. Section 3 is devoted to some unary and composition operations on coverings and the preservation of approximations under these operators. In Section 4, we introduce the concept of homomorphism and investigate the preservation properties of the lower and upper approximations under homomorphism. We conclude the paper in Section 5 with some interesting problems for further research.

## 2. Covering rough sets based on neighborhoods

This section consists of three subsections. In Section 2.1, we recall the definitions of Pawlak's rough sets and covering rough sets based on neighborhoods, and collect a few necessary facts to be used in later sections. Section 2.2 is devoted to providing an equivalent description of the upper approximation of covering rough sets without using neighborhoods. Applying the equivalent description, we briefly discuss the relationships between the lower and upper approximations in Section 2.3.

### *2.1. Approximations by neighborhoods*

We start by recalling some basic notions of Pawlak's rough set theory [24, 25].

Let $U$ be a finite and nonempty universal set, and let $R \subseteq U \times U$ be an equivalence relation on $U$. Denote by $U/R$ the set of all equivalence classes induced by $R$. Such equivalence classes, also called *elementary sets*, give a partition of $U$; every union of elementary sets is called a *definable set*. For any $X \subseteq U$, one can characterize $X$ by a pair of lower and upper approximations. The *lower approximation* $X_*$ of $X$ is defined as the greatest definable set contained in $X$, while the *upper approximation* $X^*$ of $X$ is defined as the least definable set containing $X$. Formally,

$$X_* = \cup\{C \in U/R \mid C \subseteq X\} \text{ and } X^* = \cup\{C \in U/R \mid C \cap X \neq \emptyset\}.$$

Clearly, the notion of partitions plays an important role in the above approximations. As an extension of partitions, coverings of the universe have been used to define the lower and upper approximations.

**Definition 1.** *Let $U$ be a finite and nonempty universal set, and $\mathscr{C} = \{C_i \mid i \in I\}$ a family of nonempty subsets of $U$. If $\bigcup_{i\in I} C_i = U$, then $\mathscr{C}$ is called a* covering *of $U$. The ordered pair $\langle U, \mathscr{C}\rangle$ is said to be a* covering approximation space.

It follows from the above definition that any partition of $U$ is certainly a covering of $U$. For convenience, the members of a general covering (not necessarily a partition) are also called *elementary sets*, and any union of elementary sets is called a *definable set*. In the literature, there are several kinds of rough sets induced by a covering [2, 3, 28, 38, 45, 54, 55, 57, 59]. For our purpose, we only recall the covering rough sets based on the following concept of neighborhoods [54].

**Definition 2.** *Let $\langle U, \mathscr{C}\rangle$ be a covering approximation space. For any $x \in U$, the* neighborhood *of $x$ is defined by $\mathcal{N}(x) = \bigcap\{C \in \mathscr{C} \mid x \in C\}$.*

In other words, the neighborhood of $x$ is the intersection of elementary sets containing $x$. Based on this notion, Zhu proposed the following approximations in [54].

**Definition 3.** *Let $\langle U, \mathscr{C}\rangle$ be a covering approximation space. For any $X \subseteq U$, the* lower approximation *of $X$ is defined as*

$$X^-_{\mathscr{C}} = \cup\{C \in \mathscr{C} \mid C \subseteq X\}$$

*and the* upper approximation *of $X$ is defined as*

$$X^+_{\mathscr{C}} = \cup\{\mathcal{N}(x) \mid x \in X\backslash X^-_{\mathscr{C}}\} \cup X^-_{\mathscr{C}},$$

*in which we use "$\backslash$" as set difference.*

For simplicity, we omit the subscript $\mathscr{C}$ in $X^-_{\mathscr{C}}$ and $X^+_{\mathscr{C}}$ whenever the context is clear. We will refer to "$-$" and "$+$" as the operations of obtaining the lower and upper approximations, respectively. Notice that in Definition 3 the lower approximation is the same as those in the other types of covering rough sets [28, 52, 57, 59], but the upper approximation is completely different. If the covering $\mathscr{C}$ is a partition of $U$, then it follows immediately from definition that $X^- = X_*$ and $X^+ = X^*$.

For subsequent need, let us record an example.

**Example 1.** *Let $U = \{a, b, c, d\}$, $C_1 = \{a, b\}$, $C_2 = \{a, c\}$, and $C_3 = \{b, d\}$. Then $\mathscr{C} = \{C_1, C_2, C_3\}$ is a covering of $U$. We see that $\mathcal{N}(a) = \{a\}$, $\mathcal{N}(b) = \{b\}$, $\mathcal{N}(c) = \{a, c\}$, and $\mathcal{N}(d) = \{b, d\}$. For $X = \{a, d\}$, we have that $X^- = \emptyset$ and $X^+ = \{a, b, d\}$ by Definition 3.*

The following is a characterization of the upper approximation $X^+$ due to Zhu in [54].

**Lemma 1** ([54], Theorem 1)**.** *Let $\langle U, \mathscr{C}\rangle$ be a covering approximation space. Then $X^+ = \bigcup_{x\in X} \mathcal{N}(x)$ for any $X \subseteq U$.*

### *2.2. An equivalent characterization of the upper approximation*

The purpose of this subsection is to provide an equivalent characterization of the upper approximation without using neighborhoods. To this end, we need the notion of subcovering.

**Definition 4.** *Let $\langle U, \mathscr{C}\rangle$ be a covering approximation space. Given $X \subseteq U$ and $\mathscr{C}' \subseteq \mathscr{C}$, if $\bigcup_{C\in\mathscr{C}'} C \supseteq X$, then $\mathscr{C}'$ is said to be a* subcovering *of $X$.*

By definition, a subcovering of $X$ is nothing else than a collection of elements of $\mathscr{C}$ that covers $X$. Denote by $\mathscr{C}(X)$ the set of all subcoverings of $X$. By abusing notation we may view $\mathscr{C}$ as a mapping from $\mathscr{P}(U)$ to $\mathscr{P}(\mathscr{C})$ that maps $X$ to $\mathscr{C}(X)$, where we write $\mathscr{P}(S)$ for the power set of a set $S$. Evidently, any covering $\mathscr{C}$ of the universal set $U$ can be seen as a trivial subcovering of $X \subseteq U$. For instance, in Example 1 both $\mathscr{C}' = \{C_1, C_3\}$ and $\mathscr{C}'' = \{C_2, C_3\}$ are nontrivial subcoverings of the set $X = \{a, d\}$. These, together with $\mathscr{C}$, are all the subcoverings of $X$, namely, $\mathscr{C}(X) = \{\mathscr{C}, \mathscr{C}', \mathscr{C}''\}$.

Observe that in Definition 3 and Lemma 1, only the upper approximation is dependent on the notion of neighborhoods, which seems asymmetric. With a little surprise, the following result shows us that the same upper approximation can be defined without using neighborhoods. Roughly speaking, the upper approximation of $X$ is just the intersection of all subcoverings of $X$. Of course, it does not mean that the notion of neighborhoods is useless; instead, neighborhood sometimes provides a very good characterization of the local properties of elements.

**Theorem 1.** *Suppose that $\langle U, \mathscr{C}\rangle$ is a covering approximation space. Then for any $X \subseteq U$,*

$$X^+ = \bigcap_{\mathscr{C}'\in\mathscr{C}(X)} \bigcup_{C\in\mathscr{C}'} C.$$

Proof. To prove the equality, we only need to verify that $\bigcup_{x\in X} \mathcal{N}(x) = \bigcap_{\mathscr{C}'\in\mathscr{C}(X)} \bigcup_{C\in\mathscr{C}'} C$ by Lemma 1.

Let us first show that $\bigcup_{x\in X} \mathcal{N}(x) \subseteq \bigcap_{\mathscr{C}'\in\mathscr{C}(X)} \bigcup_{C\in\mathscr{C}'} C$. Let $y \in \bigcup_{x\in X} \mathcal{N}(x)$. Then there exists some $x_y \in X$ such that $y \in \mathcal{N}(x_y)$, namely, $y \in \cap\{C \in \mathscr{C} \mid x_y \in C\}$. This means that for any elementary set $C$ containing $x_y$, we always have that $y \in C$. On the other hand, for any subcovering $\mathscr{C}'$ of $X$ there is some $C' \in \mathscr{C}'$ such that $x_y \in C'$. By the previous argument, we see that $y \in C'$, and thus $y \in \bigcup_{C\in\mathscr{C}'} C$. As the subcovering $\mathscr{C}' \in \mathscr{C}(X)$ was arbitrary, we have that $y \in \bigcap_{\mathscr{C}'\in\mathscr{C}(X)} \bigcup_{C\in\mathscr{C}'} C$. Hence, $\bigcup_{x\in X} \mathcal{N}(x) \subseteq \bigcap_{\mathscr{C}'\in\mathscr{C}(X)} \bigcup_{C\in\mathscr{C}'} C$.

Conversely, suppose that $y \in \bigcap_{\mathscr{C}'\in\mathscr{C}(X)} \bigcup_{C\in\mathscr{C}'} C$. Then we have that $y \in \bigcup_{C\in\mathscr{C}'} C$ for any $\mathscr{C}' \in \mathscr{C}(X)$. It means that for any $\mathscr{C}' \in \mathscr{C}(X)$, there exists $C' \in \mathscr{C}'$ such that $y \in C'$. Seeking a contradiction, assume that $y \notin \bigcup_{x\in X} \mathcal{N}(x)$. It implies that $y \notin \mathcal{N}(x)$ for every $x \in X$. Since $\mathcal{N}(x) = \bigcap\{C \in \mathscr{C} \mid x \in C\}$ by definition, for each $x \in X$ there is $C_x \in \mathscr{C}$ such that $x \in C_x$, but $y \notin C_x$. Let $\mathscr{C}''$ be the collection of such $C_x$'s, i.e.,

$$\mathscr{C}'' = \{C_x \in \mathscr{C} \mid \exists x \in X \text{ such that } x \in C_x, y \notin C_x\}.$$

As a result, we find that $\mathscr{C}''$ is a covering of $X$. It follows from the previous argument that there exists $C'' \in \mathscr{C}''$ such that $y \in C''$, which contradicts the construction of $\mathscr{C}''$. Therefore, $y \in \bigcup_{x\in X} \mathcal{N}(x)$, and thus $\bigcap_{\mathscr{C}'\in\mathscr{C}(X)} \bigcup_{C\in\mathscr{C}'} C \subseteq \bigcup_{x\in X} \mathcal{N}(x)$, completing the proof of the theorem.

Notice that for any $X \subseteq U$, the covering $\mathscr{C}$ of $U$ is a trivial subcovering of $X$, so we have the following corollary, which leaves out of account the covering $\mathscr{C}$.

**Corollary 1.** *Suppose that $\langle U, \mathscr{C}\rangle$ is a covering approximation space. For any $X \subseteq U$, if $\mathscr{C}(X)$ has more than one element, then*

$$X^+ = \bigcap_{\mathscr{C}'\in\mathscr{C}(X)\setminus\{\mathscr{C}\}} \bigcup_{C\in\mathscr{C}'} C.$$

Proof. By Theorem 1, we have that

$$\begin{aligned} X^+ &= \bigcap_{\mathscr{C}'\in\mathscr{C}(X)} \bigcup_{C\in\mathscr{C}'} C \\ &= \Big(\bigcup_{C\in\mathscr{C}} C\Big) \bigcap \Big(\bigcap_{\mathscr{C}'\in\mathscr{C}(X)\setminus\{\mathscr{C}\}} \bigcup_{C\in\mathscr{C}'} C\Big) \\ &= X \bigcap \Big(\bigcap_{\mathscr{C}'\in\mathscr{C}(X)\setminus\{\mathscr{C}\}} \bigcup_{C\in\mathscr{C}'} C\Big) \\ &= \bigcap_{\mathscr{C}'\in\mathscr{C}(X)\setminus\{\mathscr{C}\}} \bigcup_{C\in\mathscr{C}'} C, \end{aligned}$$

as desired.

Let us calculate an upper approximation by using the above corollary.

**Example 2.** *We revisit Example 1, where $U = \{a, b, c, d\}$, $C_1 = \{a, b\}$, $C_2 = \{a, c\}$, $C_3 = \{b, d\}$, and $\mathscr{C} = \{C_1, C_2, C_3\}$ is a covering of $U$. For $X = \{a, d\}$, we have obtained that $\mathscr{C}(X) = \{\mathscr{C}, \mathscr{C}', \mathscr{C}''\}$, in which $\mathscr{C}' = \{C_1, C_3\}$ and $\mathscr{C}'' = \{C_2, C_3\}$. It follows from Corollary 1 that*

$$\begin{aligned} X^+ &= \bigcap_{\mathscr{C}'\in\mathscr{C}(X)\setminus\{\mathscr{C}\}} \bigcup_{C\in\mathscr{C}'} C \\ &= \Big(\bigcup_{C\in\mathscr{C}'} C\Big) \bigcap \Big(\bigcup_{C\in\mathscr{C}''} C\Big) \\ &= (C_1 \cup C_3) \cap (C_2 \cup C_3) \\ &= \{a, b, d\} \cap \{a, b, c, d\} \\ &= \{a, b, d\}. \end{aligned}$$

*This is consistent with the result obtained by Definition 3 or Lemma 1.*

**Remark 1.** As we have seen, Theorem 1 provides an equivalent definition of the upper approximation based on neighborhoods. Consequently, there are two different ways to obtain $X^+$: One is to compute the neighborhood of every element of $X$ which is a "bottom-up" approach, and the other is to compute all subcoverings of $X$ which is a "top-down" approach. They approach the same problem from different perspectives and in general, we cannot conclude which one is much easier to use. In terms of manually handling the computation, if $X$ has fewer elements it seems better to use the approach based on neighborhoods, and otherwise the approach based on subcoverings may be much easier to use.

### *2.3. Relationships between the lower and upper approximations*

In [54], Zhu pointed out that the lower and upper approximations in the covering rough sets based on neighborhoods are not independent. Roughly speaking, the lower approximation operation dominates the upper one, but the converse does not hold. The following proposition was given in [54]; the proof there is based upon a series of intermediate results, so we provide a direct proof by Theorem 1.

**Proposition 1** ([54], Theorem 8)**.** *Suppose that $U$ is a universal set and $\mathscr{C}$ and $\mathscr{C}'$ are two coverings of $U$. If $X^-_{\mathscr{C}} = X^-_{\mathscr{C}'}$ for all $X \subseteq U$, then $X^+_{\mathscr{C}} = X^+_{\mathscr{C}'}$ for all $X \subseteq U$.*

Proof. Suppose, by contradiction, that there exists $X \subseteq U$ such that $X^+_{\mathscr{C}} \neq X^+_{\mathscr{C}'}$. Without loss of generality, we assume that there is a $y \in X^+_{\mathscr{C}}$, but $y \notin X^+_{\mathscr{C}'}$. Clearly, $y \notin X$, since $X \subseteq X^+_{\mathscr{C}}$ and $X \subseteq X^+_{\mathscr{C}'}$. Since $y \in X^+_{\mathscr{C}}$, we see by Theorem 1 that $y \in \bigcup_{C\in\mathscr{C}_i} C$ for any $\mathscr{C}_i \in \mathscr{C}(X)$. It forces that there is some $x_y \in X$ such that for any $C \in \mathscr{C}$, $x_y \in C$ implies $y \in C$. Otherwise, we can obtain a subcovering of $X$ such that $y$ does not belong to each member of the subcovering, a contradiction. On the other hand, as $y \notin X^+_{\mathscr{C}'}$, by Theorem 1 there is $\mathscr{C}'_0 \in \mathscr{C}'(X)$ such that $y \notin \bigcup_{C'\in\mathscr{C}'_0} C'$. Because $x_y \in X \subseteq \bigcup_{C'\in\mathscr{C}'_0} C'$, there exists $C'_0 \in \mathscr{C}'_0 \subseteq \mathscr{C}'$ such that $x_y \in C'_0$, but $y \notin C'_0$. We thus have that ${C'_0}^-_{\mathscr{C}} = {C'_0}^-_{\mathscr{C}'} = C'_0$ by the condition given in the proposition. Consequently, $C'_0$ is the union of some sets in $\mathscr{C}$, say, $C'_0 = \bigcup_{i\in I} C_i$. Then there exists $j \in I$ such that $x_y \in C_j \subseteq C'_0$. This yields that $y \in C_j$ by the previous argument that $x_y \in C$ implies $y \in C$, for any $C \in \mathscr{C}$. It contradicts the fact that $y \notin C'_0$. As a result, $X^+_{\mathscr{C}} = X^+_{\mathscr{C}'}$ for all $X \subseteq U$, finishing the proof.

As mentioned above, the converse of Proposition 1 does not hold; the reader may refer to [54] for a counterexample. It should be stressed that if $X^-_{\mathscr{C}} = X^-_{\mathscr{C}'}$ for some (not all) $X \subseteq U$, then we cannot get $X^+_{\mathscr{C}} = X^+_{\mathscr{C}'}$ in general. Nevertheless, we have the following useful observation.

**Corollary 2.** *Let $U$ be a universal set, and $\mathscr{C}$ and $\mathscr{C}'$ two coverings of $U$. If $C^-_{\mathscr{C}} = C^-_{\mathscr{C}'}$ for every $C \in \mathscr{C} \cup \mathscr{C}'$, then for any $X \subseteq U$, $X^-_{\mathscr{C}} = X^-_{\mathscr{C}'}$ and $X^+_{\mathscr{C}} = X^+_{\mathscr{C}'}$.*

Proof. By Proposition 1, we only need to show that $X^-_{\mathscr{C}} = X^-_{\mathscr{C}'}$ for any $X \subseteq U$. To this end, let $X \subseteq U$ and assume that $X^-_{\mathscr{C}} = \bigcup_{i\in I} C_i$ for some $C_i \in \mathscr{C}$ satisfying $C_i \subseteq X$. By condition, we have that $C_i = {C_i}^-_{\mathscr{C}} = {C_i}^-_{\mathscr{C}'} = \bigcup_{j\in J_i} C'_{ij}$ for some $C'_{ij} \in \mathscr{C}'$ satisfying $C'_{ij} \subseteq X$. Therefore, $X^-_{\mathscr{C}} = \bigcup_{i\in I} C_i = \bigcup_{i\in I} \bigcup_{j\in J_i} C'_{ij} \subseteq X^-_{\mathscr{C}'}$, namely, $X^-_{\mathscr{C}} \subseteq X^-_{\mathscr{C}'}$. The converse inclusion can be proven similarly. We thus obtain that $X^-_{\mathscr{C}} = X^-_{\mathscr{C}'}$, as desired.

The above corollary shows us that two coverings of a universal set give the same lower (and also upper) approximations if and only if every elementary set in a covering is a definable set (i.e., the union of some elementary sets) in the other covering, and vice versa. This implies that two coverings lead to the same approximations if and only if their elementary sets that are not a union of other elementary sets are the same. To formally state it, let us recall a concept introduced in [52, 57].

**Definition 5.** *Let $\langle U, \mathscr{C}\rangle$ be a covering approximation space. If $C \in \mathscr{C}$ cannot be written as a union of some sets in $\mathscr{C}\backslash\{C\}$, then $C$ is called* irreducible *in $\mathscr{C}$, otherwise $C$ is called* reducible.

If all reducible elements are deleted from a covering $\mathscr{C}$, the remainder is still a covering and this new covering does not have any reducible element. We call this new covering the *reduct* of the original covering and denote it as *reduct*($\mathscr{C}$).

It follows from the definition above that any set in $\mathscr{C}$ is a definable set in *reduct*($\mathscr{C}$). Using Corollary 2, we present another proof of an important theorem appearing in [52, 54, 57].

**Corollary 3.** *Let $U$ be a universal set, and $\mathscr{C}$ and $\mathscr{C}'$ two coverings of $U$. Then, $X^-_{\mathscr{C}} = X^-_{\mathscr{C}'}$ holds for all $X \subseteq U$ if and only if reduct($\mathscr{C}$) = reduct($\mathscr{C}'$).*

Proof. If *reduct*$(\mathscr{C})$ = *reduct*$(\mathscr{C}')$, then it is easy to check that $C^-_{\mathscr{C}} = C^-_{\mathscr{C}'}$ for every $C \in \mathscr{C} \cup \mathscr{C}'$. We thus get by Corollary 2 that $X^-_{\mathscr{C}} = X^-_{\mathscr{C}'}$ for all $X \subseteq U$, and hence the sufficiency holds.

Next, to see the necessity, suppose that $X^-_{\mathscr{C}} = X^-_{\mathscr{C}'}$ holds for all $X \subseteq U$. In particular, we see that $C^-_{\mathscr{C}} = C^-_{\mathscr{C}'}$ for any $C \in \mathscr{C} \cup \mathscr{C}'$, which means that any set in $\mathscr{C}$ is a union of some sets in $\mathscr{C}'$ and also any set in $\mathscr{C}'$ is a union of some sets in $\mathscr{C}$. Consequently, for any $C \in$ *reduct*$(\mathscr{C})$, we have that $C = \bigcup_{i \in I} C'_i$ for some $C'_i \in$ *reduct*$(\mathscr{C}')$. On the other hand, we also have that $C'_i = \bigcup_{j \in J} C_{ij}$ for some $C_{ij} \in$ *reduct*$(\mathscr{C})$, and thus, $C = \bigcup_{i \in I} \bigcup_{j \in J} C_{ij}$. This forces that $|I| = |J| = 1$ (writing "$|S|$" for the cardinality of a set $S$) since $C \in$ *reduct*$(\mathscr{C})$. Therefore, $C = C'_i \in$ *reduct*$(\mathscr{C}')$, namely, *reduct*$(\mathscr{C}) \subseteq$ *reduct*$(\mathscr{C}')$. The converse inclusion may be proven in a similar way. This completes the proof.

Let us illustrate the above corollary by a simple example.

**Example 3.** *Let $U = \{x_1, x_2, x_3, x_4\}$, $C_{ij} = \{x_i, x_j\}$ with $1 \leq i < j \leq 4$, and $C_{ijk} = \{x_i, x_j, x_k\}$ with $1 \leq i < j < k \leq 4$. Taking $\mathscr{C} = \{C_{ij} \mid 1 \leq i < j \leq 4\} \cup \{C_{123}, C_{124}\}$ and $\mathscr{C}' = \{C_{ij} \mid 1 \leq i < j \leq 4\} \cup \{C_{123}, C_{134}, C_{234}\}$, it is easy to see that reduct$(\mathscr{C})$ = reduct$(\mathscr{C}') = \{C_{ij} \mid 1 \leq i < j \leq 4\}$, so the coverings $\mathscr{C}$ and $\mathscr{C}'$ give the same lower (and also upper) approximations.*

Evidently, all definable sets of a covering constitute a new covering, but such a covering does not change the lower and upper approximations.

**Corollary 4.** *Suppose that $\langle U, \mathscr{C} \rangle$ is a covering approximation space. Let $\mathscr{C}_u = \{\bigcup_{C \in \mathscr{C}'} C \mid \emptyset \neq \mathscr{C}' \subseteq \mathscr{C}\}$. Then for all $X \subseteq U$, $X^-_{\mathscr{C}} = X^-_{\mathscr{C}_u}$ and $X^+_{\mathscr{C}} = X^+_{\mathscr{C}_u}$.*

Proof. It is easy to see that *reduct*$(\mathscr{C})$ = *reduct*$(\mathscr{C}_u)$. Hence, the corollary holds by Corollary 3 and Proposition 1.

## 3. Operations on coverings

In order to facilitate the computation of coverings for covering rough sets, we introduce two operations that allow us to combine, or compose, two or more coverings, as well as several operations on a single covering to modify appropriately the elements of the covering. Some properties of the lower and upper approximations preserved by the operations are also examined in this section.

For any universal set $U$, we write Cov$(U)$ for the set of all coverings of $U$. It is well-known that the number of possible coverings for a set $U$ of $n$ elements is

$$|\mathrm{Cov}(U)| = \frac{1}{2}\sum_{k=0}^{n}(-1)^k\binom{n}{k}2^{2^{n-k}}.$$

The first few of which are 1, 5, 109, 32297, 2147321017, . . .. This quickly growing sequence is entry A003465 of Sloane [33]. Since Cov$(U)$ contains a large number of coverings in general, it may be of interest to investigate the operations on these coverings.

### *3.1. Unary operations*

Let $U$ be a universal set. By Definition 5 we may view *reduct* as a unary operation on Cov$(U)$ that maps $\mathscr{C}$ to *reduct*$(\mathscr{C})$. Moreover, it is clear by definition that the operator *reduct* is idempotent in the sense that *reduct*(*reduct*$(\mathscr{C})$) = *reduct*$(\mathscr{C})$ for any $\mathscr{C} \in$ Cov$(U)$. It turns out by Corollary 3 that both $\mathscr{C}$ and *reduct*$(\mathscr{C})$ give rise to the same lower (and also upper) approximations for every subset of $U$; see also [52, 54, 57]. In other words, both the lower and upper approximations are preserved by the operator *reduct*. In this subsection, we introduce two more unary operations on Cov$(U)$ that preserve the upper (not necessarily the lower) approximations only.

Recall that $C \in \mathscr{C}$ is called irreducible if it cannot be written as a union of some sets in $\mathscr{C} \backslash \{C\}$. Oppositely, when considering intersection operation, we have the following notion of non-intersectional elementary sets.

**Definition 6.** *Let $\langle U, \mathscr{C} \rangle$ be a covering approximation space. If $C \in \mathscr{C}$ cannot be written as an intersection of some sets in $\mathscr{C} \backslash \{C\}$, then $C$ is called* non-intersectional *in $\mathscr{C}$, otherwise $C$ is called* intersectional. *Denote by int$(\mathscr{C})$ the set of all non-intersectional elementary sets in $\mathscr{C}$.*

It follows immediately from Definition 6 that $int(\mathscr{C}) \in \mathrm{Cov}(U)$ for any $\mathscr{C} \in \mathrm{Cov}(U)$. Let us illustrate the definition by an example.

**Example 4.** *Let $U = \{x_1, x_2, x_3, x_4\}$, $C_i = \{x_i\}$ with $1 \le i \le 4$, $C_{ij} = \{x_i, x_j\}$ with $1 \le i < j \le 4$, and $C_{ijk} = \{x_i, x_j, x_k\}$ with $1 \le i < j < k \le 4$. Taking $\mathscr{C} = \{C_1, C_2, C_{12}, C_{13}, C_{123}, C_{124}, C_{134}, C_{234}\}$, we see that $\mathscr{C}$ is a covering of $U$. Observe that $C_{123}$, $C_{124}$, $C_{134}$, and $C_{234}$ are non-intersectional, while $C_1$, $C_2$, $C_{12}$, and $C_{13}$ are intersectional in $\mathscr{C}$. We thus have that $int(\mathscr{C}) = \{C_{123}, C_{124}, C_{134}, C_{234}\}$, which is still a covering of $U$.*

Notice that the function $int\colon \mathrm{Cov}(U) \longrightarrow \mathrm{Cov}(U)$ that maps $\mathscr{C}$ to $int(\mathscr{C})$ is well-defined. Hence, we may view *int* as a unary operator on $\mathrm{Cov}(U)$. Clearly, by definition the operator *int* is idempotent in the sense that $int(int(\mathscr{C})) = int(\mathscr{C})$ for any $\mathscr{C} \in \mathrm{Cov}(U)$. Furthermore, we will show that the operator preserves the upper approximations. To this end, it is convenient to have the following lemma.

**Lemma 2.** *Let $\langle U, \mathscr{C}\rangle$ be a covering approximation space. Then $\mathcal{N}_{\mathscr{C}}(x) = \mathcal{N}_{int(\mathscr{C})}(x)$ for any $x \in U$, where $\mathcal{N}_{\mathscr{C}}(x)$ and $\mathcal{N}_{int(\mathscr{C})}(x)$ are the neighborhoods of $x$ with respect to $\mathscr{C}$ and $int(\mathscr{C})$, respectively.*

Proof. It follows directly from Definitions 2 and 6.

The next theorem shows us that the operator *int* preserves the upper approximations.

**Theorem 2.** *Let $\langle U, \mathscr{C}\rangle$ be a covering approximation space. Then $X^-_{int(\mathscr{C})} \subseteq X^-_{\mathscr{C}}$ and $X^+_{int(\mathscr{C})} = X^+_{\mathscr{C}}$ for all $X \subseteq U$.*

Proof. The first part follows directly from definition. For the second part, let $X \subseteq U$. Then we have by Lemma 1 that $X^+_{int(\mathscr{C})} = \bigcup_{x\in X} \mathcal{N}_{int(\mathscr{C})}(x)$ and $X^+_{\mathscr{C}} = \bigcup_{x\in X} \mathcal{N}_{\mathscr{C}}(x)$. It follows from Lemma 2 that $\mathcal{N}_{\mathscr{C}}(x) = \mathcal{N}_{int(\mathscr{C})}(x)$ for any $x \in U$. Therefore, $X^+_{int(\mathscr{C})} = X^+_{\mathscr{C}}$, as desired.

**Remark 2.** It should be pointed out that the inclusion $X^-_{int(\mathscr{C})} \subseteq X^-_{\mathscr{C}}$ may be strict, which means that *int* cannot preserve the lower approximations in general. For instance, consider Example 4. It is easy to see that ${C_1}^-_{\mathscr{C}} = C_1$ and ${C_1}^-_{int(\mathscr{C})} = \emptyset$; the former has the latter as a proper subset.

Similar to Corollary 2, we also have the following theorem. It shows us that the upper approximation operation associated with a covering is determined by the upper approximations of elementary sets.

**Theorem 3.** *Let $U$ be a universal set and $\mathscr{C}, \mathscr{C}' \in \mathrm{Cov}(U)$. If $C^+_{\mathscr{C}} = C^+_{\mathscr{C}'}$ for every $C \in \mathscr{C} \cup \mathscr{C}'$, then $X^+_{\mathscr{C}} = X^+_{\mathscr{C}'}$ for all $X \subseteq U$.*

Proof. By contradiction, assume that there exists $X \subseteq U$ such that $X^+_{\mathscr{C}} \neq X^+_{\mathscr{C}'}$. Then we have by Lemma 1 that $X^+_{\mathscr{C}} = \bigcup_{x\in X} \mathcal{N}_{\mathscr{C}}(x)$ and $X^+_{\mathscr{C}'} = \bigcup_{x\in X} \mathcal{N}_{\mathscr{C}'}(x)$, where $\mathcal{N}_{\mathscr{C}}(x) = \bigcap_{i\in I} C_i$ and $\mathcal{N}_{\mathscr{C}'}(x) = \bigcap_{j\in J} C'_j$ are the neighborhoods of $x$ with respect to $\mathscr{C}$ and $\mathscr{C}'$, respectively. Thus, there exists $x \in X$ such that $\mathcal{N}_{\mathscr{C}}(x) \neq \mathcal{N}_{\mathscr{C}'}(x)$; otherwise, $X^+_{\mathscr{C}} = X^+_{\mathscr{C}'}$. Without loss of generality, we assume that there is a $y \in \mathcal{N}_{\mathscr{C}}(x) = \bigcap_{i\in I} C_i$, but $y \notin \mathcal{N}_{\mathscr{C}'}(x) = \bigcap_{j\in J} C'_j$. Evidently, $y \neq x$ and there is $j' \in J$ such that $y \notin C'_{j'}$. On the other hand, we find that $y \in C'_{j'}$ because $C'_{j'} = {C'_{j'}}^+_{\mathscr{C}'} = {C'_{j'}}^+_{\mathscr{C}} = \bigcup_{z\in C'_{j'}} \mathcal{N}_{\mathscr{C}}(z) \supseteq \mathcal{N}_{\mathscr{C}}(x)$ by condition and $y \in \mathcal{N}_{\mathscr{C}}(x)$. It is a contradiction. Hence, $X^+_{\mathscr{C}} = X^+_{\mathscr{C}'}$ for all $X \subseteq U$, as desired.

Note that both *reduct* and *int* are operators on $\mathrm{Cov}(U)$. It is interesting to consider their compositions. Let us write the composition of operators from right to left.

**Remark 3.** We now check the compositions of *reduct* and *int*. We find that $reduct \circ int \neq int \circ reduct$ in general. For instance, consider the covering $\mathscr{C} = \{C_1, C_2, C_{12}, C_{13}, C_{123}, C_{124}, C_{134}, C_{234}\}$ in Example 4. There is no difficulty to get that $reduct \circ int = \{C_{123}, C_{124}, C_{134}, C_{234}\}$ and $int \circ reduct = \{C_2, C_{13}, C_{124}, C_{134}, C_{234}\}$; they are different. Nevertheless, we have by Corollary 2 and Theorem 3 that $X^+_{\mathscr{C}} = X^+_{reduct\circ int(\mathscr{C})} = X^+_{int\circ reduct(\mathscr{C})}$ for all $X \subseteq U$.

Let us end this subsection with a brief discussion on the so-called neighborhood operator. We remark that a notion similar to neighborhood operator, called induced covering, was defined in [35] for another type of covering rough sets. Let $U$ be a universal set. For any $\mathscr{C} \in \mathrm{Cov}(U)$, define $nei(\mathscr{C}) = \{\mathcal{N}(x) \mid x \in U\}$. In other words, *nei* maps every covering to the set of all neighborhoods (with respect to the covering) of elements of $U$. Clearly, the set of all neighborhoods gives rise to a covering of $U$. Hence, we have that $nei(\mathscr{C}) \in \mathrm{Cov}(U)$ and thus *nei* yields a unary operator on $\mathrm{Cov}(U)$, called *neighborhood operator*.

Like the operators *reduct* and *int*, the neighborhood operator *nei* preserves the upper approximations as well.

**Theorem 4.** *Let $\langle U, \mathscr{C}\rangle$ be a covering approximation space. Then $X^-_{\mathscr{C}} \subseteq X^-_{nei(\mathscr{C})}$ and $X^+_{\mathscr{C}} = X^+_{nei(\mathscr{C})}$ for all $X \subseteq U$.*

Proof. For the first part, note that $\bigcup_{x\in X^-_{\mathscr{C}}} \mathcal{N}(x) \supseteq X^-_{\mathscr{C}}$ and $X \supseteq \mathcal{N}(x)$ for any $x \in X^-_{\mathscr{C}}$. We thus have that

$$\begin{aligned} X^-_{nei(\mathscr{C})} &= \cup\{\mathcal{N}(x) \mid x \in X \text{ such that } \mathcal{N}(x) \subseteq X\} \\ &\supseteq \cup_{x\in X^-_{\mathscr{C}}} \mathcal{N}(x) \\ &\supseteq X^-_{\mathscr{C}}, \end{aligned}$$

namely, $X^-_{\mathscr{C}} \subseteq X^-_{nei(\mathscr{C})}$.

The second part follows immediately from Lemma 2.

**Remark 4.** We remark that the inclusion $X^-_{\mathscr{C}} \subseteq X^-_{nei(\mathscr{C})}$ may be strict. For example, setting $U = \{a, b, c\}$ and $\mathscr{C} = \big\{\{a, b\}, \{b, c\}, \{a, c\}\big\}$, we get that $nei(\mathscr{C}) = \big\{\{a\}, \{b\}, \{c\}\big\}$. Taking $X = \{a\}$, we see that $X^-_{\mathscr{C}} = \emptyset$ and $X^-_{nei(\mathscr{C})} = X$; the former is properly included in the latter.

*3.2. Composition operations*

In this subsection, we address the following problems: For a given universal set $U$, if there are two coverings $\mathscr{C}_1$ and $\mathscr{C}_2$ of $U$, can we construct a new covering of $U$ via $\mathscr{C}_1$ and $\mathscr{C}_2$? Further, if we get a new covering of $U$, what are the relationships between the upper (lower) approximations with respect to the new covering and the original coverings? To this end, we define two operations on coverings: the union, denoted by $\vee$, and the intersection, denoted by $\wedge$. For simplicity, we present these operations for two coverings.

Let us begin with the union operation.

**Definition 7.** *Let $U$ be a universal set and $\mathscr{C}_1, \mathscr{C}_2 \in Cov(U)$. The* union *of $\mathscr{C}_1$ and $\mathscr{C}_2$, denoted by $\mathscr{C}_1 \vee \mathscr{C}_2$, is defined as $\mathscr{C}_1 \cup \mathscr{C}_2$, the usual union of sets $\mathscr{C}_1$ and $\mathscr{C}_2$.*

In other words, the union operation is to collect all elementary sets in each covering. Clearly, $\mathscr{C}_1 \vee \mathscr{C}_2 \in \mathrm{Cov}(U)$ whenever $\mathscr{C}_1, \mathscr{C}_2 \in \mathrm{Cov}(U)$. Further, we have the following property.

**Theorem 5.** *Let $U$ be a universal set and $\mathscr{C}_1, \mathscr{C}_2 \in Cov(U)$. Then for all $X \subseteq U$, $X^-_{\mathscr{C}_i} \subseteq X^-_{\mathscr{C}_1\vee\mathscr{C}_2}$ and $X^+_{\mathscr{C}_1\vee\mathscr{C}_2} \subseteq X^+_{\mathscr{C}_i}$, where $i = 1, 2$.*

Proof. Let $X \subseteq U$. Then by definition we see that

$$\begin{aligned} X^-_{\mathscr{C}_1\vee\mathscr{C}_2} &= \cup\{C \in \mathscr{C}_1 \cup \mathscr{C}_2 \mid C \subseteq X\} \\ &= (\cup\{C \in \mathscr{C}_1 \mid C \subseteq X\}) \cup (\cup\{C \in \mathscr{C}_2 \mid C \subseteq X\}) \\ &\supseteq \cup\{C \in \mathscr{C}_i \mid C \subseteq X\} \\ &= X^-_{\mathscr{C}_i}, \end{aligned}$$

namely, $X^-_{\mathscr{C}_i} \subseteq X^-_{\mathscr{C}_1\vee\mathscr{C}_2}$.

For the second part, it follows from the fact $\mathscr{C}_i \subseteq \mathscr{C}_1 \vee \mathscr{C}_2$ and Theorem 1 that

$$\begin{aligned} X^+_{\mathscr{C}_1\vee\mathscr{C}_2} &= \bigcap_{\mathscr{C}\in\mathscr{C}_1\vee\mathscr{C}_2(X)} \bigcup_{C\in\mathscr{C}} C \\ &\subseteq \bigcap_{\mathscr{C}\in\mathscr{C}_i(X)} \bigcup_{C\in\mathscr{C}} C \\ &= X^+_{\mathscr{C}_i}, \end{aligned}$$

that is, $X^+_{\mathscr{C}_1\vee\mathscr{C}_2} \subseteq X^+_{\mathscr{C}_i}$, as desired. This completes the proof of the theorem.

We now turn our attention to the intersection operation.

**Definition 8.** *Let $U$ be a universal set and $\mathscr{C}_1, \mathscr{C}_2 \in Cov(U)$. The* intersection *of $\mathscr{C}_1$ and $\mathscr{C}_2$, denoted by $\mathscr{C}_1 \wedge \mathscr{C}_2$, is defined as $\{\mathcal{N}_{\mathscr{C}_1\cup\mathscr{C}_2}(x) \mid x \in U\}$, where $\mathcal{N}_{\mathscr{C}_1\cup\mathscr{C}_2}(x) = \cap\{C \in \mathscr{C}_1 \cup \mathscr{C}_2 \mid x \in C\}$.*

The intersection of $\mathscr{C}_1$ and $\mathscr{C}_2$ is nothing else than the set of neighborhoods of all elements of the universal set. It should be noted that the neighborhoods are defined with respect to the union of $\mathscr{C}_1$ and $\mathscr{C}_2$ and by set-theoretic intersection; hence the term intersection. Again, $\mathscr{C}_1 \wedge \mathscr{C}_2 \in \mathrm{Cov}(U)$ whenever $\mathscr{C}_1, \mathscr{C}_2 \in \mathrm{Cov}(U)$. Like Theorem 5, we have the following property.

**Theorem 6.** *Let $U$ be a universal set and $\mathscr{C}_1, \mathscr{C}_2 \in Cov(U)$. Then for all $X \subseteq U$, $X^-_{\mathscr{C}_i} \subseteq X^-_{\mathscr{C}_1\wedge\mathscr{C}_2}$ and $X^+_{\mathscr{C}_1\wedge\mathscr{C}_2} \subseteq X^+_{\mathscr{C}_i}$, where $i = 1, 2$.*

Proof. For any $x \in U$, let $\mathcal{N}(x)$ denote the neighborhood of $x$ with respect to the covering $\mathscr{C}_1 \wedge \mathscr{C}_2$. Then by the construction of $\mathscr{C}_1 \wedge \mathscr{C}_2$ we always have that $\mathcal{N}(x) = \mathcal{N}_{\mathscr{C}_1\cup\mathscr{C}_2}(x)$. In addition, for any $C \in \mathscr{C}_1 \cup \mathscr{C}_2$ and $x \in C$, we see that $\mathcal{N}(x) \subseteq C$. Therefore, $\bigcup_{x\in C} \mathcal{N}(x) \subseteq C$. Clearly, $C \subseteq \bigcup_{x\in C} \mathcal{N}(x)$, and we thus obtain that $C = \bigcup_{x\in C} \mathcal{N}(x)$, i.e., $C = \bigcup_{x\in C} \mathcal{N}_{\mathscr{C}_1\cup\mathscr{C}_2}(x)$. It means that every elementary set in $\mathscr{C}_i$ $(i = 1, 2)$ is a union of some elementary sets in $\mathscr{C}_1 \wedge \mathscr{C}_2$. This forces that $X^-_{\mathscr{C}_i} \subseteq X^-_{\mathscr{C}_1\wedge\mathscr{C}_2}$ for all $X \subseteq U$.

For the second part, note that we always have that $\mathcal{N}_{\mathscr{C}_1\cup\mathscr{C}_2}(x) \subseteq \mathcal{N}_{\mathscr{C}_i}(x)$, $i = 1, 2$, for any $x \in U$, that is, $\mathcal{N}(x) \subseteq \mathcal{N}_{\mathscr{C}_i}(x)$. It follows immediately from Lemma 1 that $X^+_{\mathscr{C}_1\wedge\mathscr{C}_2} \subseteq X^+_{\mathscr{C}_i}$, finishing the proof.

In fact, the union and intersection operations of coverings are related, as shown below.

**Proposition 2.** *Let $U$ be a universal set and $\mathscr{C}_1, \mathscr{C}_2 \in Cov(U)$. Then $\mathscr{C}_1 \wedge \mathscr{C}_2 = nei(\mathscr{C}_1 \vee \mathscr{C}_2)$.*

Proof. It follows directly from Definitions 7 and 8 and the definition of neighborhood operator.

The next example illustrates the composition operations defined above.

**Example 5.** *As in Example 4, let $U = \{x_1, x_2, x_3, x_4\}$, $C_i = \{x_i\}$ with $1 \leq i \leq 4$, $C_{ij} = \{x_i, x_j\}$ with $1 \leq i < j \leq 4$, and $C_{ijk} = \{x_i, x_j, x_k\}$ with $1 \leq i < j < k \leq 4$. Take $\mathscr{C}_1 = \{C_{12}, C_{24}, C_{234}\}$ and $\mathscr{C}_2 = \{C_{123}, C_{234}\}$. Then by definition we get that $\mathscr{C}_1 \vee \mathscr{C}_2 = \{C_{12}, C_{24}, C_{123}, C_{234}\}$ and $\mathscr{C}_1 \wedge \mathscr{C}_2 = \{C_{12}, C_2, C_{23}, C_{24}\}$. For $X = C_{23}$, we can readily obtain by a routine computation that $X^-_{\mathscr{C}_1} = X^-_{\mathscr{C}_2} = X^-_{\mathscr{C}_1\vee\mathscr{C}_2} = \emptyset$, $X^-_{\mathscr{C}_1\wedge\mathscr{C}_2} = X^+_{\mathscr{C}_2} = X^+_{\mathscr{C}_1\vee\mathscr{C}_2} = X^+_{\mathscr{C}_1\wedge\mathscr{C}_2} = C_{23}$, and $X^+_{\mathscr{C}_1} = C_{234}$.*

## 4. Homomorphisms between covering approximation spaces

In this section, we look at the preservation properties of the lower and upper approximations under homomorphism, a mapping between covering approximation spaces. The concept of homomorphism makes it possible to relate different coverings to different agents or moments in time.

**Definition 9.** *Let $\langle U, \mathscr{C}\rangle$ and $\langle V, \mathscr{D}\rangle$ be two covering approximation spaces. A mapping $f : U \longrightarrow V$ is called a* homomorphism *from $\langle U, \mathscr{C}\rangle$ to $\langle V, \mathscr{D}\rangle$ if it maps each element of $\mathscr{C}$ to an element in $\mathscr{D}$.*

Recall that by elementary sets we mean the members of a covering. The unions of elementary sets are referred to as definable sets. Clearly, the above definition is equivalent to say that $f : U \longrightarrow V$ is a homomorphism if it maps each definable set of $\langle U, \mathscr{C}\rangle$ to a definable set of $\langle V, \mathscr{D}\rangle$. Note that the above definition of homomorphism for covering approximation spaces is an extended definition of the homomorphism for information systems [7, 10, 15, 36, 46, 51].

Let us consider several examples of homomorphism between covering approximation spaces.

**Example 6.** *Let $\langle U, \mathscr{C}\rangle$ be a covering approximation space. Denote by $id_U$ the identity mapping on $U$ that maps every element to itself. Then the following facts hold:*

*(1) $id_U$ is both a homomorphism from $\langle U, \mathscr{C}\rangle$ to $\langle U, reduct(\mathscr{C})\rangle$ and a homomorphism from $\langle U, reduct(\mathscr{C})\rangle$ to $\langle U, \mathscr{C}\rangle$.*

*(2) $id_U$ is a homomorphism from $\langle U, \mathscr{C}\rangle$ to $\langle U, nei(\mathscr{C})\rangle$, but not a homomorphism from $\langle U, nei(\mathscr{C})\rangle$ to $\langle U, \mathscr{C}\rangle$ in general.*

*(3) $id_U$ is a homomorphism from $\langle U, int(\mathscr{C})\rangle$ to $\langle U, \mathscr{C}\rangle$, but not a homomorphism from $\langle U, \mathscr{C}\rangle$ to $\langle U, int(\mathscr{C})\rangle$ in general.*

Homomorphisms have the following property.

**Lemma 3.** *If $f : \langle U, \mathscr{C}\rangle \longrightarrow \langle V, \mathscr{D}\rangle$ is a homomorphism between covering approximation spaces, then $f(X^-) \subseteq f(X)^-$ for any $X \subseteq U$.*

Proof. Given $X \subseteq U$, assume that $X^- = \bigcup_{i\in I} C_i$ for some $C_i \in \mathscr{C}$ satisfying $C_i \subseteq X$, and also assume that $f(C_i) = \bigcup_{j_i\in J_i} D_{j_i}$ for some $D_{j_i} \in \mathscr{D}$. Evidently, $D_{j_i} \subseteq f(X)$ for all $i \in I$ and $j_i \in J_i$. We thus have that

$$\begin{aligned} f(X^-) &= f(\cup_{i\in I} C_i) \\ &= \cup_{i\in I} f(C_i) \\ &= \cup_{i\in I} \cup_{j_i\in J_i} D_{j_i} \\ &\subseteq f(X), \end{aligned}$$

i.e., $f(X^-) \subseteq f(X)$. It means that $f(X^-) \subseteq f(X)^-$ since every $D_{j_i} \in \mathscr{D}$. This completes the proof.

One may wonder whether there is a corresponding inclusion relation between $f(X^+)$ and $f(X)^+$. As shown in the next remark, the answer is "no", in general.

**Remark 5.** Let $U = \{x_1, x_2, x_3, x_4, x_5\}$, $V = \{y_1, y_2, y_3, y_4\}$, $\mathscr{C} = \{\{x_1, x_2\}, \{x_2, x_3\}, \{x_4, x_5\}\}$, and $\mathscr{D} = \{\{y_1, y_2\}, \{y_3\}, \{y_4\}\}$. Then we get two covering approximation spaces $\langle U, \mathscr{C}\rangle$ and $\langle V, \mathscr{D}\rangle$. Setting $f(x_1) = f(x_3) = y_1$, $f(x_2) = y_2$, $f(x_4) = y_3$, and $f(x_5) = y_4$ and taking $X = \{x_2, x_4\}$, we obtain that $X^+ = \{x_2, x_4, x_5\}$ and $f(X) = \{y_2, y_3\}$. Further, we have that $f(X^+) = \{y_2, y_3, y_4\}$ and $f(X)^+ = \{y_1, y_2, y_3\}$. As we see, there is no inclusion relation between $f(X^+)$ and $f(X)^+$.

Recall that a mapping is said to be *bijective* if it is both injective and surjective. If a homomorphism $f$ between two covering approximation spaces is bijective, and moreover, the inverse mapping $f^{-1}$ of $f$ is also a homomorphism, then $f$ is called an *isomorphism*. For instance, the homomorphism $id_U$ from $\langle U, \mathscr{C}\rangle$ to $\langle U, reduct(\mathscr{C})\rangle$ is an isomorphism.

The following theorem shows us that isomorphisms preserve the lower and upper approximations.

**Theorem 7.** *If $f : \langle U, \mathscr{C}\rangle \longrightarrow \langle V, \mathscr{D}\rangle$ is an isomorphism between covering approximation spaces, then for any $X \subseteq U$, $f(X^-) = f(X)^-$ and $f(X^+) = f(X)^+$.*

To prove the theorem, it is convenient to have the following lemma, which says that the image of the neighborhood of $x$ under an isomorphism $f$ is exactly the neighborhood of $f(x)$.

**Lemma 4.** *If $f : \langle U, \mathscr{C}\rangle \longrightarrow \langle V, \mathscr{D}\rangle$ is an isomorphism between covering approximation spaces, then for any $x \in U$, $f(\mathcal{N}_{\mathscr{C}}(x)) = \mathcal{N}_{\mathscr{D}}(f(x))$.*

Proof. For any $x \in U$, suppose that $\mathcal{N}_{\mathscr{C}}(x) = \bigcap_{i\in I} C_i$ for some $C_i \in \mathscr{C}$ with $x \in C_i$, and suppose that $f(C_i) = \bigcup_{j_i\in J_i} D_{j_i}$ for some $D_{j_i} \in \mathscr{D}$. Note that $x \in C_i$, hence $f(x) \in f(C_i)$, and thus, there exists $0_i \in J_i$ such that $f(x) \in D_{0_i}$. Because $f$ is isomorphic, we get that

$$\begin{aligned} f(\mathcal{N}_{\mathscr{C}}(x)) &= f(\cap_{i\in I} C_i) \\ &= \cap_{i\in I} f(C_i) \\ &= \cap_{i\in I} \cup_{j_i\in J_i} D_{j_i} \\ &\supseteq \cap_{i\in I} D_{0_i} \\ &\supseteq \mathcal{N}_{\mathscr{D}}(f(x)), \end{aligned}$$

namely, $f(\mathcal{N}_{\mathscr{C}}(x)) \supseteq \mathcal{N}_{\mathscr{D}}(f(x))$. Since $f^{-1}$ is a homomorphism, we have that $f^{-1}[\mathcal{N}_{\mathscr{D}}(f(x))] \supseteq \mathcal{N}_{\mathscr{C}}(f^{-1}(f(x))) = \mathcal{N}_{\mathscr{C}}(x)$, i.e., $f^{-1}[\mathcal{N}_{\mathscr{D}}(f(x))] \supseteq \mathcal{N}_{\mathscr{C}}(x)$. Applying $f$ to the inclusion, we see that $\mathcal{N}_{\mathscr{D}}(f(x)) \supseteq f(\mathcal{N}_{\mathscr{C}}(x))$. This, together with the previous argument, forces that $f(\mathcal{N}_{\mathscr{C}}(x)) = \mathcal{N}_{\mathscr{D}}(f(x))$, thus proving the lemma.

We can now present a proof of Theorem 7.

Proof of Theorem 7. Let $X \subseteq U$. We first verify that $f(X^-) = f(X)^-$. In fact, it follows at once from Lemma 3 that $f(X^-) \subseteq f(X)^-$. Applying $f^{-1}$ to the inclusion, we see that $f^{-1}[f(X^-)] \subseteq f^{-1}[f(X)^-]$, namely, $X^- \subseteq f^{-1}[f(X)^-]$. On the other hand, we have by Lemma 3 that $f^{-1}[f(X)^-] \subseteq f^{-1}[f(X)]^- = X^-$ since $f^{-1}$ is a homomorphism and $f(X) \subseteq V$. Consequently, $X^- \subseteq f^{-1}[f(X)^-] \subseteq X^-$, which forces that $X^- = f^{-1}[f(X)^-]$. Applying $f$ to the equality gives rise to $f(X^-) = f(X)^-$, as desired.

Let us turn now to the proof of $f(X^+) = f(X)^+$. Using Lemma 4 and the condition that $f$ is bijective, we obtain by Lemma 1 that

$$\begin{aligned} f(X^+) &= f(\cup_{x\in X} \mathcal{N}_{\mathscr{C}}(x)) \\ &= \cup_{x\in X} f(\mathcal{N}_{\mathscr{C}}(x)) \\ &= \cup_{x\in X} \mathcal{N}_{\mathscr{D}}(f(x)) \\ &= \cup_{f(x)\in f(X)} \mathcal{N}_{\mathscr{D}}(f(x)) \\ &= \cup_{y\in f(X)} \mathcal{N}_{\mathscr{D}}(y) \\ &= f(X)^+, \end{aligned}$$

that is, $f(X^+) = f(X)^+$. This completes the proof of the theorem.

As mentioned above, the homomorphism $id_U : \langle U, \mathscr{C}\rangle \longrightarrow \langle U, reduct(\mathscr{C})\rangle$ is an isomorphism. Therefore, the following result given in [52, 54, 57] is a direct corollary of Theorem 7.

**Corollary 5.** *Let $\langle U, \mathscr{C}\rangle$ be a covering approximation space. Then for any $X \subseteq U$, $X^-_{\mathscr{C}} = X^-_{reduct(\mathscr{C})}$ and $X^+_{\mathscr{C}} = X^+_{reduct(\mathscr{C})}$.*

## 5. Conclusion

In this paper, we have explored more properties of the covering rough sets based on neighborhoods. It has been shown that the upper approximation based on neighborhoods can be defined equivalently without using the notion of neighborhoods. Several operations on coverings and homomorphisms between covering approximation spaces have been introduced to covering rough sets. We have also verified the properties of the lower and upper approximations preserved by the operations and homomorphisms, respectively. Broadly speaking, after providing the equivalent definition of the upper approximation based on neighborhoods, we have focused on the preservation property of the lower and upper approximations under different coverings. In particular, the unary operators *int* and *nei* make it possible to simplify the computation of upper approximations by preprocessing a covering; the composition operations $\vee$ and $\wedge$ are helpful to estimate the lower and upper approximations in some cases; the notion of homomorphism, especially isomorphism, builds a bridge between two covering approximation spaces which makes upper (lower) approximations comparable under the homomorphism.

The present work is mainly concerned with the covering rough sets based on neighborhoods. It would be interesting to examine the operations on coverings and homomorphisms between covering approximation spaces for other types of covering rough sets. In addition, some other issues in topology such as continuous maps and homeomorphisms remain yet to be addressed in the covering rough sets based on neighborhoods. In fact, recall that it was shown by McKinsey and Tarski [21, 22] that if we interpret modal diamond as the closure (or equivalently, the box is interpreted as the interior) in a topological space, then the modal logic of topological spaces is exactly Lewis' well-known modal system S4. It has been known that open and continuous maps (called interior maps) preserve modal validity. What we call a homomorphism in the paper is actually an open map between coverings whereas an isomorphism is an interior map between coverings. In light of this, one may investigate further coverings with some links to topology.

## Acknowledgements

This work was supported by the National Natural Science Foundation of China under Grants 60821001, 60873191, 60903152, and 61070251. The author would like to thank Editor-in-Chief, Area Editor, and the anonymous referees for some valuable comments and helpful suggestions.